\documentclass[10pt]{article}
\usepackage[letterpaper]{geometry}
\usepackage{hicss}
\usepackage{times}
\usepackage[none]{hyphenat}
\usepackage{url}
\usepackage{latexsym}
\usepackage{indentfirst}
\usepackage{graphicx}
\usepackage{cite}
\usepackage{amsmath}
\usepackage{amssymb}
\usepackage{amsthm}
\usepackage{algorithm}
\usepackage[noend]{algpseudocode}
\usepackage{enumitem}
\usepackage{fancybox}
\usepackage{tcolorbox}
\usepackage{multirow}
\usepackage{pifont}
\usepackage{xcolor}

\definecolor{niceblue}{rgb}{0, 0.5, 1.0}
\graphicspath{{images/}}
\usepackage{float}
\floatstyle{ruled}

\newfloat{model}{thp}{lop}
\floatname{model}{Model}

\newfloat{problem}{thp}{lop}
\floatname{problem}{Problem}

\newtheorem{lemma}{Lemma}
\newtheorem{remark}{Remark}

\title{GPU-Accelerated Verification of Machine Learning Models \\for Power Systems}

 \author{Samuel Chevalier, Ilgiz Murzakhanov, and Spyros Chatzivasileiadis\\
 Department of Wind and Energy Systems \\
  Lyngby, Denmark \\
  {{ \{schev, ilgmu, spchatz\}@dtu.dk}} \\
}

\date{}

\begin{document}
\maketitle
\begin{abstract}
Computational tools for rigorously verifying the performance of large-scale machine learning (ML) models have progressed significantly in recent years. The most successful solvers employ highly specialized, GPU-accelerated branch and bound routines. Such tools are crucial for the successful deployment of machine learning applications in safety-critical systems, such as power systems. Despite their successes, however, barriers prevent out-of-the-box application of these routines to power system problems. This paper addresses this issue in three key ways. {First, we reformulate several key power system verification problems into the canonical format utilized by modern verification solvers.} Second, we enable the simultaneous verification of multiple verification problems (e.g., checking for the violation of all constraints simultaneously, and not by solving individual verification problems). 
To achieve this, we introduce an exact transformation that converts a set of potential violations into a series of ReLU-based neural network layers. This allows verifiers to interpret these layers directly, and determine the ``worst-case'' violation in a single shot. 
Third, power system ML models often must be verified to satisfy power flow constraints. 
We propose a dualization procedure which encodes linear equality and inequality constraints (such as power balance constraints and line flow constraints) directly into the verification problem in a manner which is mathematically consistent with the specialized verification tools. To demonstrate these innovations, we verify problems associated with data-driven security constrained DC-OPF solvers. We build and test our first set of innovations using the $\alpha,\beta$-CROWN solver, and we benchmark against Gurobi 10.0. Our contributions achieve a speedup that can exceed 100x and allow higher degrees of verification flexibility.
\end{abstract}

\subsubsection*{Keywords:}

Branch and bound, data driven modeling, DC-OPF, neural network verification, machine learning.

\section{Introduction}
The ubiquity of machine learning (ML) applications has led to a surge of interest in topics related to adversarial robustness~\cite{carlini2023certified}, performance verification~\cite{Wang:2021}, and safety guarantees~\cite{Fazlyab:2022,Venzke:2020}. Collectively, these tools allow ML users to have a higher degree of \textit{trust} in the underlying black-box models developed by learning algorithms. In order to spur the development and implementation of such tools, the EU Commission recently proposed the creation of regulations which will help assess ML risk, engender certification requirements, and enforce industry standards~\cite{kop2021eu}. The National Institute of Standard and Technologies (NIST) in the US is looking to develop similar technical standards to help with ML regulation~\cite{NIST_AI,national2019us}.

In this paper, we focus on the problem of performance verification, i.e., formally verifying that a Neural Network (NN) model obeys certain input-output mapping properties. In particular, we focus on performance verification within the context of network-constrained, safety critical applications (i.e., electrical power system operation). Performance verification tools have recently been developed for data-driven models used in a number of power system applications. Within the reinforcement learning (RL) context, authors in~\cite{hosseini2022verification} developed analytical feasibility ellipsoids where RL solutions were guaranteed to be feasible. This work was extended in~\cite{Hosseini:2023}, where non-feasible RL solutions were projected into
convex polyhedron feasibility sets. Other works have exploited the mixed integer linear programming (MILP) NN reformulation proposed in~\cite{Tjeng:2017}. In~\cite{Venzke:2020_fg}, NN verification procedures were introduced for the first time for power system classification problems: NNs were trained to classify the stability of power system operating points, and adversarial inpyts or the resulting classification areas were certified though the MILP reformulation of NNs. Taking advantage of the underlying knowledge of the physical models, a nover verification procedure was introduced in~\cite{Venzke:2020}, where it was now possible to determine not just if there is a violation, but rather the extent of the worst-case violation. The method was applied for NNs estimating the solutions of linear programs (DC-OPF) in power systems. The method was subsequenty extended to determine the worst-case performance of NN-based AC-OPF solvers (i.e. non-linear programs) in~\cite{NELLIKKATH2022108412}; the resulting optimization problems were shown to be extremely hard to solve (requiring up to 5 hours of compute time for small power systems). Inspired by the challenges in~\cite{NELLIKKATH2022108412}, authors in~\cite{chevalier2023global} used semidefinite programming and Sherali-Adam cut generation to rapidly generate worst-case performance bounds for NN models of AC power flow.

Despite its usefulness, NN performance verification is a very challenging computational problem. Encouragingly, the ML research community itself has made rapid progress on developing \textit{efficient} methods for formally verifying NN behavior. The International Verification of Neural Networks Competition (VNN-COMP)~\cite{müller2023international,bak2021second}, which is currently on its fourth iteration, has attracted significant research attention and helped spur many fantastic innovations. The results of these competitions, which are summarized in~\cite{brix2023years}, have shown that highly specialized Branch-and-Bound (BaB) solvers, combined with rapid bound propagation approaches~\cite{zhang2018efficient,xu2020automatic}, have come to dominate other approaches based on, e.g., semidefinite or quadratic programming~\cite{Dathathri:2020,Krishnamurthy:2018,Fazlyab:2022}. Two of the most successful verification solvers are $\alpha,\beta$-CROWN~\cite{Wang:2021} and Multi-Neuron Constraint Guided BaB (MN-BAB)~\cite{ferrari2022complete} (in this paper, we refer to these tools as ``verifiers"). Both tools have the ability to formally verify NNs containing tens of thousands of neurons, and they can be efficiently implemented on GPUs.

While tools such as $\alpha,\beta$-CROWN~\cite{Wang:2021} are enormously powerful, they currently cannot be directly applied to many of the power system verification problems described in the previous paragraphs. In this paper, we help to bridge this gap by proposing three key innovations. These innovations, and their associated challenges, are summarized below.

\begin{enumerate}
   
    \item In contrast to verifying for a single safety metric each time, power system verification problems need to be able to verify across many hundreds or thousands of safety metrics (e.g., across all individual line flow violations at the same time). This paper introduces, for the first time to our knowledge, a NN extension (i.e., a series of ReLU-based layers) which \textit{exactly} computes the ``worst case" violation across some set of potential violations. The reformulated verification problem can then be directly interpreted by verification solvers and solved in a single shot.
    
    \item Power system verification problems must often contain physical constraints such as network equation constraints. Unfortunately, there is no way to naturally incorporate such constraints into the verification solvers which have emerged from VNN-COMP. Consistent with $\alpha,\beta$-CROWN, we develop a custom dualization procedure which incorporates linear equality and inequality constraints into the Lagrange dual objective function. This allows a verifier to solve network-constrained verification problems.

    \item Finally, the verification standards which have emerged from the ML community can be challenging to interpret and apply. Accordingly, before elaborating on our first two key contributions, we first introduce a procedure to reformulate power system verification problems into the canonical format utilized by modern verification solvers. 
\end{enumerate}
The remainder of this paper is organized as follows. In Section \ref{sec: Complete_Verification}, we review the problem of complete verification of NNs, and we provide several power system applications. In Section \ref{sec: Adapting_BaB}, we explain the challenges associated with directly applying specialized complete verifiers to power system problems, and we offer several solutions for overcoming these challenges. Test results are given in Section \ref{sec: Test_results}, and conclusions are drawn in Section \ref{sec: Conclusion}

\section{Complete Verification of NNs}\label{sec: Complete_Verification}
\begin{figure}
\centering
\includegraphics[width=0.95\linewidth]{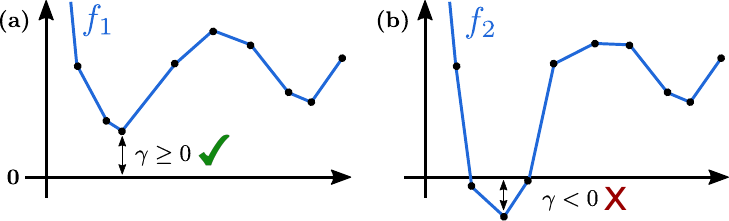}
	\caption{Verification of NN performance metrics.}
	\label{fig: nnv}
\end{figure}

Neural network (NN) performance verification seeks to prove that a NN mapping $y={\rm NN}(x)$ satisfies a given performance metric applied to the output, denoted here as $m(y)$. By convention, if this metric is proved to be positive across the NN's full input domain, then the model has been \textit{verified}\cite{wang2022efficient}. Otherwise, the desired performance metric has been \textit{refuted}. To state this problem mathematically, we define $f(x)$ as the composition of the performance metric and the NN mapping: $f(x) \triangleq m({\rm NN}(x))$. Next, we define an optimization problem which searches for the minimum value of $f(x)$ across the NN input domain\cite{wang2022efficient}:
\begin{align}\label{eq: nnv_gamma}
\gamma=\min_{x\in\mathcal{C}}\; & f(x),
\end{align}
{were $\mathcal{C}$ is an $\ell_p$ norm ball representing the allowable input region~\cite{wang2022efficient,Wang:2021}.} The sign of $\gamma$ indicates if the NN's performance has been either verified or refuted:
\begin{subequations}\label{eq: gam}
\begin{align}
\gamma\ge0 &\;\Rightarrow\; \text{metric verified (pass)}\label{eq: gam_pass}\\
\gamma<0 &\;\Rightarrow\; \text{metric refuted (fail)}.
\end{align}
\end{subequations}
Both conditions are depicted in Figure~\ref{fig: nnv} for a piece-wise linear NN. Panel $\bf{(a)}$ shows a successful verification with $\gamma\ge 0$, and panel $\bf{(b)}$ shows a unsuccessful verification $\gamma<0$.

Verification solvers which can provide a definitive solution to \eqref{eq: nnv_gamma} given sufficient computational time are referred to as \textit{complete} verifiers~\cite{wang2022efficient}. Incomplete verifiers, on the other hand, typically search for adversarial inputs or use sophisticated bound propagation in an attempt to solve \eqref{eq: nnv_gamma}, sometimes unsuccessfully. Due to the classification of power systems as safety-critical infrastructure, the verification tools applied to ML problems in this field will need to be strong enough to prove any given performance metric stated by energy regulators or other stakeholders. Accordingly, in this paper, we focus on the the problem of \textit{complete} verification.

\subsection{Application to power: verification of learning-based DC-OPF solvers}
In this paper, we focus on verifying various performance metrics associated with NNs which directly predict DC-OPF solutions\footnote{While DC-OPF is the guiding application to demonstrate our methods in this paper, our contributions can be applied to a plethora of MILP-related verification problems.}. Consider a power system with associated load vector $p_{d}\in {\mathbb R}^n$ and generation vector $p_{g}\in {\mathbb R}^n$. Many recent works have used NNs to directly learn the mapping from a load set-point to an ``$N-0$" optimal generator dispatch (i.e., a dispatch which satisfies all generation limits and line flow limits for the intact system):
\begin{align}
p_{g} = {\rm NN}_{\text{n-0}}(p_{d}).
\end{align}
In order to ensure that no individual generator is ever dispatched above its respective upper bound (for a given bound on the loads), we may pose the verification problem stated in Problem \ref{prob1}, which is consistent with \eqref{eq: nnv_gamma}. In this model, reformulation \eqref{eq: non-convex_reformulation} from Appendix A1 is employed to find the worst case violation across all potential violations. Notably, if $t\ge 0$ across all possible load variations in $\mathcal{C}$, then $\overline p_{g}\ge{p}_{g}$ and no violation occurs.
Constraint \eqref{eq: NN} ensures that the NN generator dispatch corresponds to the given load. Constraint \eqref{eq: ubsub} ensures that any generator limit violation is captured by the auxiliary variable $t$, large constant $M$, and binary variable $b$ of \eqref{eq: b}. Constraint \eqref{eq: sum} is a binary restriction which ensures that only the worst-case generation limit is captured.
\begin{problem}
\caption{\hspace{-0.1cm}\textbf{:} Verification of DC-OPF Generator Limits}
\label{prob1}
\vspace{-0.25cm}
\begin{subequations}
\begin{align}
\gamma=\min_{p_{d}\in\mathcal{C}}\quad & t\\
{\rm s.t.}\quad & {\rm NN}_{\text{n-0}}(p_{d})=p_{g} \label{eq: NN}\\
 & \overline{p}_{g,i}-p_{g,i}\le t+Mb_{i},\;\forall i\in\mathcal{N}\label{eq: ubsub}\\
 & \sum b_i=n-1\label{eq: sum}\\
 & b\in\{0,1\}^{n}.\label{eq: b}
\end{align}
\end{subequations}
\vspace{-0.5cm}
\end{problem}

We now consider a secondary problem which seeks to verify if an ``N-1" DC-OPF solution actually satisfies network flow limits. To state this problem, we introduce incidence matrix $E\in {\mathbb R}^{m\times n}$, diagonal susceptance matrix $Y_{b}$, nodal phase angle vector $\theta\in {\mathbb R}^{n}$, upper and lower line flow limits $\overline{p}_{f},\underline{p}{}_f\in\mathbb{R}^{m}$, binaries $b^{{\rm on}}\in\mathbb{Z}^{m}$ which indicate if a line is in service or not, and binaries $b^{{\rm u}},b^{{\rm l}}\in\mathbb{Z}^{m}$ which capture the worst case error. Finally, ${\rm NN}_{\text{n-1}}$ is trained to compute N-1 secure DC-OPF solutions. The verification problem is formally stated in Problem \ref{prob2}; the formulation is similar in form to, e.g., ~\cite{Han:2021}. Constraints \eqref{eq: Mflow} and \eqref{eq: Mb} capture the flow conditions under the N-1 criterion. Constraint \eqref{eq: Eflow} represents the power balance condition at each node in the network. Constraints \eqref{eq: bU} and \eqref{eq: bL} represent the upper and lower limits on the line flow, respectively. The constraints \eqref{eq: b2m} and \eqref{eq: bm} enforce restrictions on the number of line flow violations and the minimum number of lines that must be in service, respectively. 
\begin{problem}
\caption{\hspace{-0.1cm}\textbf{:} Verification of N-1 DC-OPF Flow Limits}
\label{prob2}
\vspace{-0.25cm}
\begin{subequations}
\begin{align}
\gamma=\min_{p_{d}\in\mathcal{C}}\quad & t\\
{\rm s.t.}\quad & {\rm NN}_{\text{n-1}}(p_{d})=p_{g}\\
 & -\!M(1\!-\!b^{{\rm on}})\le p_{f}-Y_{b}E\theta\nonumber \label{eq: Mflow}\\
 &\quad\quad\quad\quad\quad\quad\quad\quad \le M(1\!-\!b^{{\rm on}})\\
 & -\!Mb_{i}^{{\rm on}}\le p_{f,i}\le Mb_{i}^{{\rm on}} \label{eq: Mb}\\
 & p_{g}-p_{d}=E^{T}p_{f} \label{eq: Eflow}\\
 & \overline{p}_{f,i}-p_{f,i}\le t + b_{i}^{{\rm u}}M \label{eq: bU}\\
 & p_{f,i}-\underline{p}_{f,i}\le t + b_{i}^{{\rm l}}M\label{eq: bL}\\
 & \sum b_{i}^{{\rm u}}+b_{i}^{{\rm l}}=2m-1 \label{eq: b2m}\\
 & \sum b_{i}^{{\rm on}}\ge m-1 \label{eq: bm}\\
 & b_{i}^{{\rm u}},b_{i}^{{\rm l}},b_{i}^{{\rm on}}\in\{0,1\}. \label{eq: allb}
\end{align}
\end{subequations}
\vspace{-0.5cm}
\end{problem}

\subsection{Branch and Bound Search Routines}
As evidenced by the recent NN verification competitions~\cite{müller2023international,bak2021second}, specialized complete BaB search routines have become the most successful tool for solving \eqref{eq: nnv_gamma}. We briefly review the dualization procedure utilized by one of the top verifiers in the academic marketplace: $\alpha,\beta$-CROWN~\cite{Wang:2021}. In subsequent sections, we will build on this framework. {Notably, we could have built our innovations on top of an alternative bound-propagation-based verifier, such as MN-BAB, but we utilized $\alpha,\beta$-CROWN due to its dominance in the recent verification competitions, and the general maturity of its code base.}

Consider the simple verification problem with two bounded, scalar inputs $x_1,x_2\in{\mathbb R}$:
\begin{align}\label{eq: gam_example}
\gamma=\min_{\underline{x}\le x\le\overline{x}}\; & w_{1}\sigma(x_{1})+w_{2}\sigma(x_{2})+b\end{align}
where the goal is to prove \eqref{eq: gam}. The strategy for solving \eqref{eq: gam_example} in a complete sense has four primary steps~\cite{Wang:2021}: ($i$) bound the activation functions, ($ii$) ``branch" on subsets of ReLU activations, ($iii$) dualize resulting inequality constraints, and ($iv$) use projected gradient descent to prove a lower bound. These steps are briefly explained.

\begin{figure}
\centering
\includegraphics[width=1.00\linewidth]{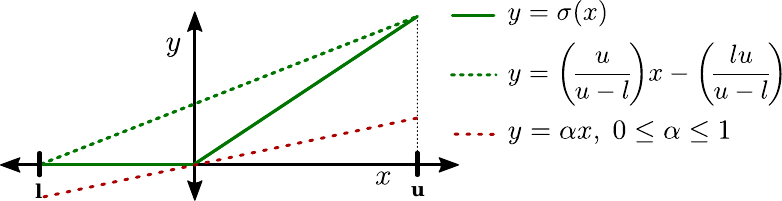}
	\caption{ReLU bounds.}
	\label{fig: ReLU_Bounds} 
\end{figure}

In step 1, all (non-fixed) ReLU activation functions are replaced by a function $g(x)$ which lower bounds \eqref{eq: gam_example}. If $w>0$, then the lower bound is given by the red dashed line in Figure \ref{fig: ReLU_Bounds}; if $w<0$, then the lower bound is given by the green dashed line:
\begin{align}
\gamma\ge\tilde{\gamma}=\min_{\underline{x}\le x\le\overline{x}}\; & g(x_{1})+g(x_{2})+b.
\end{align}
Notably, if $\tilde{\gamma}\ge 0$, then $\gamma\ge 0$.

In step 2, a subset of ReLU activation functions are branched on (i.e., their statuses are assumed known). Assuming $x_1$ is branched in the ``on" direction, we have
\begin{subequations}
\begin{align}
\gamma\ge\tilde{\gamma}=\min_{\underline{x}\le x\le\overline{x}}\; & x_{1}w_{1}+g(x_{2})+b\\
{\rm s.t.}\;\, & -x_{1}\le0.\label{eq: constr_dual}
\end{align}
\end{subequations}
In step 3, we form the Lagrangian by pushing constraint \eqref{eq: constr_dual} into the objective function and creating the dual problem~\cite{boyd2004convex}:
\begin{align}\label{eq: LD}
\gamma\ge\tilde{\gamma}=\max_{\beta\ge0}\min_{\underline{x}\le x\le\overline{x}}\; & x_{1}w_{1}+g(x_{2})+b-\beta x_{1},
\end{align}
where $\beta$ is a positive Lagrange multiplier. Notably, the inner minimization from \eqref{eq: LD} has a closed form solution via the dual norm (see Appendix  A2). Exploiting this property, \cite{Wang:2021} solves \eqref{eq: LD} by rewriting the objective function as $a(\beta)^{T}x+b$, where vector $a$ is parameterized by the dual variable $\beta$. We also assume, without loss of generality, that the input bounds can be transformed and expressed with a $p$ norm (see Appendix A3):
\begin{subequations}\label{eq: ld_solve}
\begin{align}
{\tilde \gamma} \;=\;& \max_{\beta\ge 0}\min_{\left\Vert x\right\Vert_{\boldsymbol p}\le\epsilon}\,a(\beta)^{T}x\!+\!b\\
=\;&\max_{\beta\ge 0}\,-\epsilon\left\Vert a(\beta)\right\Vert_{\boldsymbol q}\!+\!b,\label{eq: ld_solve_soln}
\end{align}
\end{subequations}
where $\boldsymbol{p}$ and $\boldsymbol{q}$ are conjugate norms, as in \eqref{eq: pq_conj}. The interested reader is referred to \cite{zhang2018efficient} for more details. 
This procedure is sequentially repeated for every layer in a given NN (i.e., bound, branch, and dualize). As a final step, the RHS of \eqref{eq: ld_solve_soln} is solved via projected gradient descent iterations\footnote{The tunable value of $0\le\alpha\le 1$ from the lower red line in Figure \ref{fig: ReLU_Bounds} is also iterated on, further tightening the bound.}. If, at any time, its value crosses above 0, then that branch of the BaB tree has been verified (i.e., \eqref{eq: gam_pass} is known for that branch and everything below it), and iterations are terminated. Backpropogation through \eqref{eq: ld_solve} can exploit GPUs, making this procedure very fast.

\section{Adapting Branch and Bound Solvers for use in Power System Applications}\label{sec: Adapting_BaB}
In this section, we offer two key adaptations which make the aforementioned specialized Branch and Bound strategies amenable to power system applications. 

\subsection{Explicitly Encoding Worst-Case Violations as NN Layers}\label{ss: encoding}
Optimizing for the ``worst of the worst", as we do in \eqref{eq: ubsub} to find the worst case generator limit violation, is a hard problem. In Appendix A1, we show how the non-convex problem $\min_{x,y\in{\mathcal C}}\;\min(x,y)$ can be reformulated using integer variables. The key observation here is that the minimization operator can also be captured using a ReLU activation function $\sigma(\cdot)$.
\begin{lemma}\label{lem: relu}
The function $\min(x,y)$ can be written with a ReLU activation function via $\min(x,y)=x-\sigma(x-y)$.
\begin{proof}
Assume $x<y$. Then $\sigma(x-y)=0$ and $\min(x,y)=x-0=x$. Assume $x>y$. Then $\sigma(x-y)=x-y$ and $\min(x,y)=x-x+y=y$.
\end{proof}
\end{lemma}
Using successive applications of this result, we may replace the mixed integer reformulation of the minimization operator with a series of NN layers. For example, consider the minimization across four variables subtracted from their upper bounds (fully analogous to \eqref{eq: ubsub}:
\begin{align}\label{eq: delta}
\delta=\min(\overline{x}_{1}-x_{1},\overline{x}_{2}-x_{2},\overline{x}_{3}-x_{3},\overline{x}_{4}-x_{4}).
\end{align}
The output $\delta$ can be equivalently computed via the composition of three NN layers ($l_1$, $l_2$, and $l_3$):
\begin{subequations}\label{eq: reform}
\begin{align}
&\left(\!\!\begin{array}{c}
\overline{x}_{1}-x_{1}\rightarrow y_{1}\\
\overline{x}_{2}-x_{2}\rightarrow y_{2}\\
\overline{x}_{3}-x_{3}\rightarrow y_{3}\\
\overline{x}_{4}-x_{4}\rightarrow y_{4}
\end{array}\!\!\right)_{l_1}  \rightarrow\\
 & \quad\quad \quad \left(\!\!\begin{array}{c}
y_{1}-\sigma(y_{1}-y_{2})\rightarrow z_{1}\\
y_{3}-\sigma(y_{3}-y_{4})\rightarrow z_{2}
\end{array}\!\!\right)_{l_2}\rightarrow\\
 & \quad \quad\quad\quad\quad \quad \quad \left(\!\begin{array}{c}
z_{1}-\sigma(z_{1}-z_{2})\end{array}\!\right)_{l_3}\rightarrow\delta.
\end{align}
\end{subequations}
Note that in \eqref{eq: reform}, it was an arbitrary choice to group $y_1$ with $y_2$ and $y_3$ with $y_4$. As the goal is to calculate the minimum across four variables, the choice of pairs is not inherent to the problem.
This reformulation allows us to rewrite Problem \ref{prob1} without any \textit{explicit} binary variables. This is highly convenient, because it translates the verification problem into a ``language" that NN verifiers can directly interact with. We note that not all mixed integer problems can be reformulated in this way: this is a special case.
\begin{remark}\label{rem: relu2n}
The minimization across $2^n$ variables, as in \eqref{eq: delta}, requires $2^n-1$ ReLU activation functions. This is because, at each new layer, each variable is paired with another variable and a ReLU is applied (one ReLU for every pairing until the very last layer).
\end{remark}
For example, the reformulation of \eqref{eq: delta} with $2^2=4$ variables required the introduction of $2^2-1=3$ ReLUs in \eqref{eq: reform}. Notably, this is the same number of binaries which must be introduced in \eqref{eq: ubsub}. Alg. \ref{algo:nn_layers} summarizes the procedure we use to encode this reformulation across an arbitrary number of NN outputs. In this algorithm, the matrices $I_1$ and $I_2$ represent selection matrices which, respectively, select the first half and the second half of a vector. For example, in the 2- dimensional case,
\begin{align*}
\underbrace{\left[\begin{array}{cc}
1 & 0\\
0 & 0
\end{array}\right]}_{I_{1}}\left[\begin{array}{c}
y_{1}\\
y_{2}
\end{array}\right]=y_{1},\quad\underbrace{\left[\begin{array}{cc}
0 & 0\\
0 & 1
\end{array}\right]}_{I_{2}}\left[\begin{array}{c}
y_{1}\\
y_{2}
\end{array}\right]=y_{2}.
\end{align*}
{Importantly, line 9 of the algorithm concatenates a subset of the layer's input ($z_1$) with a second subset (${\tilde z}$) which is explicitly transformed.}

\begin{algorithm}
\caption{Encoding Minimization as NN Layers}\label{algo:nn_layers}

{\small \textbf{Require:}
Neural network $y={\rm NN}(x)$ with $n_y$ outputs, each associated with an element of upper bound vector ${\overline y}$

\begin{algorithmic}[1]

\State Initialize: ${\hat n}_y\leftarrow n_y$, $\hat{\rm NN}\leftarrow {\rm NN}$

\State Append $\hat{\rm NN}$: ${\overline y}-y\rightarrow z$

\While{${\hat n}_y > 1$}

\If{${\hat n}_y = 2k, \quad \forall k \in \mathbb{Z}$}

\State Append $\hat{\rm NN}$ layer: $z \leftarrow I_1 z - \sigma(I_1 z - I_2 z)$

\State ${\hat n}_y \leftarrow {\hat n}_y/2$

\Else 

\State Split: $
z_{1}\cup\tilde{z}\leftarrow z$

\State Append $\hat{\rm NN}$ layer: $z \leftarrow\left[\!\!\begin{array}{c}
z_{1}\\
I_{1}\tilde{z}-\sigma(I_{1}\tilde{z}-I_{2}\tilde{z})
\end{array}\!\!\!\!\right]$

\State ${\hat n}_y \leftarrow ({\hat n}_y-1)/2 + 1$

\EndIf {\bf end}

\EndWhile {\bf end}

\State \Return $\hat{\rm NN}$

\end{algorithmic}}
\end{algorithm}

\subsection{Extending dualization-based BaB to include general equality and inequality constraints}
In power system applications, ML models often need to be verified within the context of complex network constraints. Consider Problem \ref{prob2}. Direct verification of this model with $\alpha,\beta$-CROWN, even with the worst-case reformulation from the previous subsection, is not possible; this is for three primary reasons:
\begin{enumerate}[noitemsep]
    \item There is no mechanism for including \textit{equality} constraints, such as \eqref{eq: Eflow}, in the verification \vspace{0.25em}
    \item There is no mechanism for including \textit{inequality} constraints, such as \eqref{eq: Mb}, in the verification\vspace{0.25em}
    \item Without inequalities, branching and bounding over general binary relaxations $0\le b \le 1$ and splits (as would be needed for turning lines on and off with $b^{\rm on}$) is not possible.
\end{enumerate}

We now formulate the inclusion of these constraints in a mathematical framework which is fully consistent with $\alpha,\beta$-CROWN (i.e., formulation \eqref{eq: ld_solve}). To begin, we consider the verification problem\footnote{ Notably, \eqref{eq: dualize_og} could represent an actual verification problem, or it could represent an LP relaxed ``node" of a BaB search tree, searching over various binary splits. For example, \eqref{eq: dualize_og} can capture Problem \ref{prob2} if the subset of binaries in \eqref{eq: allb} corresponding to line status are either fixed or relaxed due to the simplification of a MILP problem into an LP problem.}:
\begin{subequations}\label{eq: dualize_og}
\begin{align}
\gamma=\min_{x,y,z}\quad & c^{T}z\label{eq: gam_LD1}\\
{\rm s.t.}\quad & y={\rm NN}(x)\\
 & Ay=z\;:\;\lambda \label{eq: dualize_og_eq}\\
 & Bz\le h\;:\; \mu,\label{eq: dualize_og_ineq}
\end{align}
\end{subequations}
where $\lambda$ and $\mu$ are associated dual variables. We now dualize both constraint sets by forming the Lagrangian $\mathcal{L}=c^{T}z+\lambda^{T}\left(Ay-z\right)+\mu^{T}(Bz-h)$, minimizing over the primals, and maximizing over the duals:
\begin{subequations}\label{eq: abc_amenable}
\begin{align*}
\gamma=\max_{\mu\ge0,\lambda}\min_{x,y,z} \quad & \lambda^{T}Ay+(c+B^{T}\mu-\lambda)^{T}z-\mu^T h\\
{\rm s.t.}\quad & y={\rm NN}(x).
\end{align*}
\end{subequations}
No relaxation has occurred through dualizing, so the values of $\gamma$ in \eqref{eq: gam_LD1} and its dualized counterpart are equivalent. Note that an extended version of \eqref{eq: dualize_og} can accommodate any linear equality and inequality constraints (e.g., without being limited to a case  where the NN output $y$ and another set of variables $z$ are related through \eqref{eq: dualize_og_eq} and \eqref{eq: dualize_og_ineq}).

The objective function in \eqref{eq: abc_amenable} can now be viewed as a generalized performance metric function $m(y,z)$ which is optimized over for the purpose of verification. Importantly, since this function is a linear function of $y$, where $y={\rm NN}(x)$, we may now apply the sequential ``bound, branch, and dualize" steps of the vanilla $\alpha,\beta$-CROWN algorithm, as previously reviewed. Using the ``$a(\beta)^{T}x+b$" term from the LHS of \eqref{eq: ld_solve} to represent NN dualization, we have
\begin{align}\label{eq: updated_abc}
\gamma\ge \tilde{\gamma}=\max_{\beta,\mu\ge0,\lambda}&\min_{\left\Vert x\right\Vert _{p}\le\epsilon,z}\; \lambda^{T}A\left(a(\beta)^{T}x+b\right)\nonumber+\\[-10pt]
& \quad \quad \, (c+B^{T}\mu-\lambda)^{T}z-\mu^T h,
\end{align}
where $\tilde{\gamma}$ is a lower bound due to ReLU relaxation. Before solving the inner minimization in \eqref{eq: updated_abc}, the bound on the primal variable $z$ must be considered. The variable $x$ naturally has a bound given by the limits of the NN verification problem, but the limits on $z$ are less clear. For example, $z$ may correspond to phase angle variables $\theta$ in Problem \ref{prob2}, and placing an a priori bound on these variables might cut off some region of the feasible space, thus nullifying the verification solution (i.e., if the verification problem is not solved over the full input region, then the results are not conclusive). Without a bound, however, the dual norm cannot be applied to analytically solve the inner minimization. Furthermore, an extremely loose bound will significantly slow down the rate of solution convergence in gradient descent. We therefore have the following remark.
\begin{remark}\label{re: 1}
    The primal bound on $z$ in \eqref{eq: updated_abc} should be maximally tight without shrinking the feasible space.
\end{remark}

Concatenating primals via $u=\{x,z\}$, we write the objective in \eqref{eq: updated_abc} as $d(\mu,\lambda,\beta)^{T}u+c(\mu,\lambda,\beta)$. Using Remark \ref{re: 1}, we bound and normalize $u$, such that $\left\Vert u\right\Vert _{p}\le\epsilon$ (see Appendix A3). We now rewrite \eqref{eq: updated_abc} as
\begin{align}\label{eq: ld_solve_soln_update}
{\tilde \gamma}=&\max_{\beta,\mu\ge0,\lambda}\min_{\left\Vert u\right\Vert _{p}\le\epsilon}d(\mu,\lambda,\beta)^{T}u+c(\mu,\lambda,\beta).
\end{align}
Solving the inner minimization of \eqref{eq: ld_solve_soln_update} with the dual norm, we have the following key result:
\begin{tcolorbox}[boxsep=1pt,left=1pt,right=4pt,top=-5pt,bottom=4pt]
\begin{equation}\label{eq: ld_solve_soln_update_final} 
{\tilde \gamma}= \max_{\beta,\mu\ge0,\lambda}-\epsilon\left\Vert d(\mu,\lambda,\beta)\right\Vert _{q}+c(\mu,\lambda,\beta).
\end{equation}
\end{tcolorbox}
Solution \eqref{eq: ld_solve_soln_update_final} is fully analogous to \eqref{eq: ld_solve_soln}; now, however, we have directly included physics-based equality and inequality constraints into the dual norm solution. Using GPU-accelerated back-propagation, \eqref{eq: ld_solve_soln_update_final} can be solved with any projected gradient descent algorithm (e.g., Adam).

The modifications needed to solve \eqref{eq: ld_solve_soln_update_final} can be added directly to the $\alpha,\beta$-CROWN solver (or any BaB solver which can solve MILPs). However, a full update of the $\alpha,\beta$-CROWN source code is beyond the scope of this paper. In the test results section, we test the validity of the bounds produced with \eqref{eq: ld_solve_soln_update_final} on a small, 4-bus example, and we validate against Gurobi's BaB solver.

\section{Test Results}\label{sec: Test_results}
In this section, we present test results associated with two verification experiments: generation limit violations, as posed by Problem \ref{prob1}, and N-1 line flow violations, as posed by Problem \ref{prob2}. In both cases, we train dense, ReLU-based NNs to predict the associated regression solutions, and then we formally verify properties of the trained NNs.

\subsection{Verification of Generator Bounds}
In the first experiment, we trained a dense NN with 10 layers and 1000 ReLU activation functions on DC-OPF data collected from the IEEE 300-bus test case \cite{uw300bus}. We then posed a verification problem which sought to identify if any set of load setpoints (bounded at $\pm25\%$ from the nominal) led to a generator limit violation, as posed by Problem \ref{prob1}. Each load could freely vary within $\pm25\%$ of its nominal value. We tested if the NN outputs for any combination of the loads setpoint resulted to a violation of any of the 68 generator limits. In order to test a wider variety of conditions, we scaled generators limits from 80\% to 120\% in 10\% increments, testing for violations in each case.

We solved this verification problem in two ways: ($i$) directly with Gurobi 10.0's MILP solver, and ($ii$) with $\alpha,\beta$-CROWN \textit{after} appending the trained NN with additional layers, via Alg.~\ref{algo:nn_layers}, to directly compute the worst case violation. The verification problems were solved on a High-Performance Computing (HPC) server with an Intel Xeon E5-2650v4 processor and an NVIDIA A100 GPU with 40 GB of RAM.

Table~\ref{table:your_label} demonstrates the numerical results for both Gurobi and $\alpha,\beta$-CROWN solver. 
The check-mark indicates verification, and the x-mark indicates violation. We observe that generator limits had to be increased by $20\%$ w.r.t. the nominal to prevent violation. Gurobi tests were constrained to a duration of 1 hour, and callbacks were used to terminate the Gurobi solver once the verification problem was solved (i.e., once $t$ crossed 0).

Several conclusions can be drawn from Table~\ref{table:your_label}. First, for problems where it was relatively easy to identify a violation, $\alpha,\beta$-CROWN with the modified NN outperformed Gurobi by solving approximately seven times faster. The problem difficulty increases as we increase the generation limits, as it is more difficult to detect generation limit violations. Eventually, a complete verifier should go through all possible branches to verify no violations. In the most challenging verification problem, with a $120\%$ generation limit, $\alpha,\beta$-CROWN returned results within 18 seconds, whereas Gurobi was unable to complete the verification assignment within the given  time constraint of 1 hour, meaning $\alpha,\beta$-CROWN solved approximately \textit{200x} faster than Gurobi. These results show that $\alpha,\beta$-CROWN is far more proficient at solving large-scale DC-OPF MILP verification problems than Gurobi is. Equally importantly, the results also demonstrate how appending layers onto pre-trained NNs allows the $\alpha,\beta$-CROWN solver to verify across many different verification metrics simultaneously. This paper is the first to introduce a technique for the verification of multiple metrics simultaneously -- and with a minimal computational overhead. This has so far not been possible with existing approaches.

\begin{table}[ht]
\centering
\caption{Computing Time Comparison of Gurobi and $\alpha,\beta$-CROWN solvers on the IEEE 300-bus test case.}
\label{table:your_label}
\begin{tabular}{|c|c|c|c|}
\hline
Gen lim & \multicolumn{2}{c|}{Gurobi} & $\alpha,\beta$-CROWN\\
\cline{2-4} 
 (\% nom) & Gap ($\%$) & Time (s) & Time (s) \\
\hline
80\% \;\ding{55} &  27 & 43.00  &  6.00 \\ 
\hline
90\% \;\ding{55} & 79 & 37.00 & 5.60 \\ 
\hline
100\% \;\ding{55} & 150  & 35.00 & 5.71 \\ 
\hline
110\% \;\ding{55} & 511 & 39.00  &  5.58 \\ 
\hline
120\% \;\ding{51} & 1726 & $>$3600 (\textbf{dnf}) & 18.03 \\ 
\hline
\end{tabular}
\end{table}

\subsection{Verification of N-1 NN Predictions}

\begin{figure}
\centering
\includegraphics[width=0.75\linewidth]{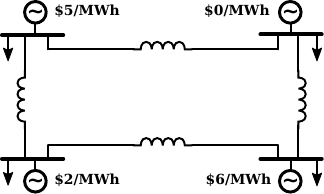}
	\caption{4-bus, 4-line power system.}
	\label{fig: 4_bus_system} 
\end{figure}

\begin{figure*}
\centering
\includegraphics[width=1.00\linewidth]{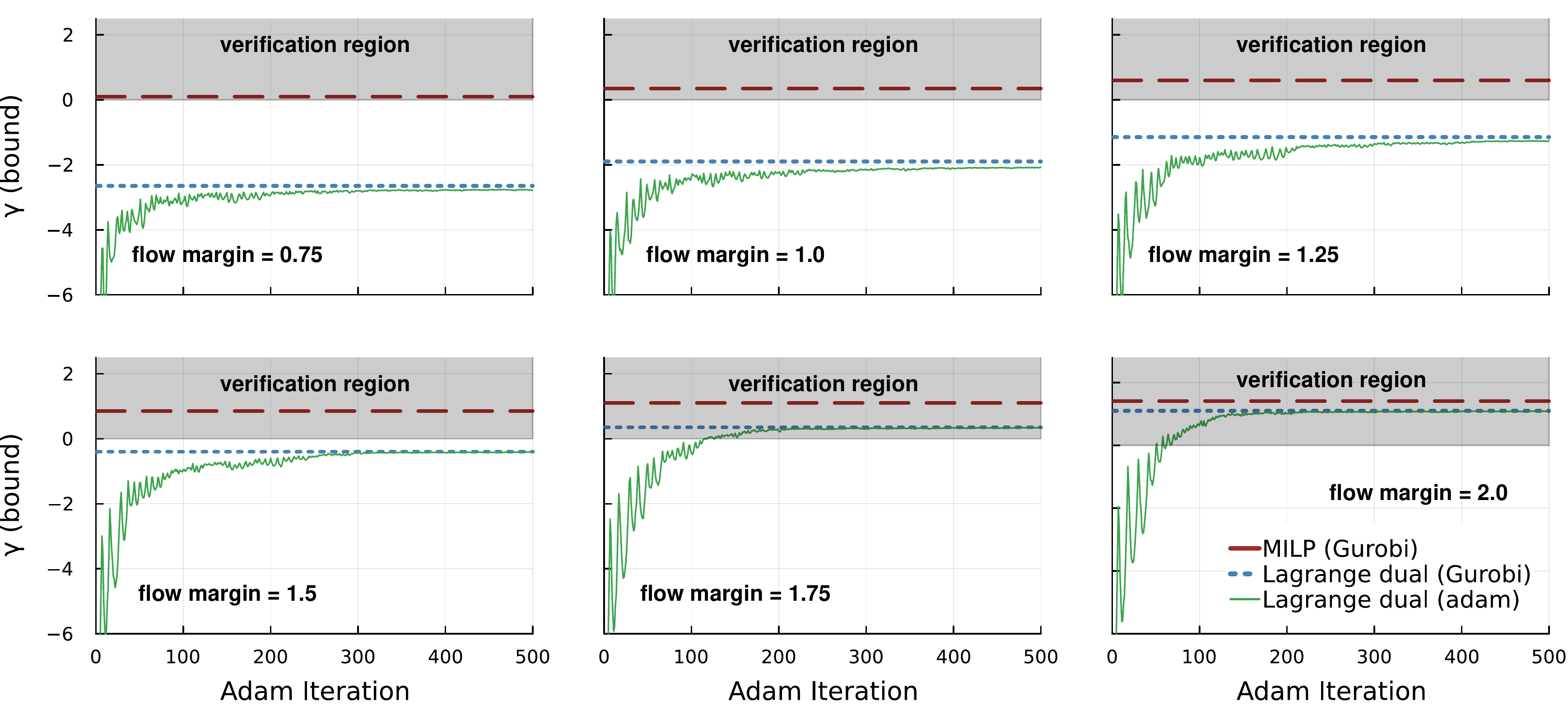}
	\caption{N-1 secure verification bounds for the 4-bus power system. {The true solution (Gurobi) is given by the red dashed curve, the root relaxation solution (Gurobi) is given by the blue curve, and the Lagrange dual, as solved by projected Adam, is given by the green solid curve.}}
	\label{fig: verification_bounds} 
\end{figure*}

In this subsection, we solve a 
small-scale version of Problem \ref{prob2} using a 4-bus, 4-line power system, as depicted in Figure~\ref{fig: 4_bus_system}. {That is, we tested if a NN which was trained to solve security constrained DC OPF would ever produce a dispatch schedule which violated an N-1 constraint.} We pose and solve this problem in order to test and validate \eqref{eq: ld_solve_soln_update_final}. In order to perform this test, we took the following steps.
\begin{enumerate}
    \item As outlined in Appendix A.4, we collected 1000 N-1 secure DC-OPF solutions from the 4-bus system under randomly sampled loads (uniformly ranging from 0 to 2pu), with rejection for infeasible samples.
    \item Using this data, we trained a small, 2-layer NN with 4 ReLU activation functions.
    \item After training, we then formulated verification Problem \ref{prob2} in order to determine if there existed some input which led to NN outputs with N-1 violation. Loads could only vary between 0 and 0.1 pu\footnote{We set a very small load range for demonstration pusposes, so that a subset of the root relaxations would verify -- this can be seen in the last two panels of Figure~\ref{fig: verification_bounds}, where the Adam iterations successfully cross into the verification region.}. We solved the MILP with Gurobi for a variety of line flow limits (i.e., increasing the line flow limits by a scaling factor, thus lowering the probability of a violation).
    \item Finally, we relaxed the MILP (i.e., root relaxation, where all binaries are relaxed) and posed \eqref{eq: ld_solve_soln_update_final}, which was solved with both Gurobi (for benchmarking) and projected Adam\footnote{The step-by-step conversion of Problem \ref{prob2} into \eqref{eq: ld_solve_soln_update_final} is not included in this manuscript due to space limitations. However, the Julia code implementation will be made available on the authors' public GitHub repository.}.
\end{enumerate}

The results of these tests are depicted in Figure~\ref{fig: verification_bounds} for various line flow margin scaling factors (i.e., we scaled the nominal line flow limits ${\overline p}_f$, ${\underline p}_f$ by some common factor). The upper dashed line represents the true, MILP solution of Problem \ref{prob2}{, as solved by Gurobi}. Notably, each subproblem passed the ground truth MILP verification test (since the red dashed lines are in the gray regions), {meaning the NN predictions are guaranteed ``safe".}

We then posed a root relaxation of Problem \ref{prob2} to generate \eqref{eq: ld_solve_soln_update_final}. This was in turn solved by \textit{both} Gurobi (dotted blue line in Figure \ref{fig: verification_bounds}) and by projected Adam (i.e., the green projected gradient descent iterations in Figure \ref{fig: verification_bounds}). In each test, the Adam iterations {correctly} converged towards the Gurobi root relaxation solution. In the first four tests (depicted in the first four panels), however, with flow margins between $0.75$ and $1.5$, the root relaxation could not provide a verification certificate: in these four cases, the lower bound never crossed 0 into the gray region. Without a zero-crossing, we cannot obtain a rigorous guarantee {about whether or not the true solution (which upper bounds the green trace) will ever cross 0 or not: no conclusion can be drawn.}

In this experiment, we only solved the root relaxation; the next step would be to fix a subset of binaries and continue down the BaB tree. The purpose of this experiment, however, was to showcase the validity of \eqref{eq: ld_solve_soln_update_final} within the context of equality and inequality constrained NN verification, which we have successfully done via Figure~\ref{fig: verification_bounds}. Gurobi, of course, can solve \eqref{eq: ld_solve_soln_update_final} directly without resorting to gradient iterations. However, the power of $\alpha,\beta$-CROWN and other specialized verifiers is that they rely heavily on GPU-based backpropagation to solve \eqref{eq: ld_solve_soln_update_final}, and they can terminate early once a zero-crossing is detected (for example, around iteration 50 in the final panel of Figure \ref{fig: verification_bounds}, {or around iteration 115 in the second-to-last panel}). Gurobi and other traditional optimization solvers, which exploit CPU-based log-barrier and simplex methods, cannot as easily exploit either of these luxuries. {While confirming the mathematical validity of \eqref{eq: ld_solve_soln_update_final} was the primary purpose for running the tests in Figure~\ref{fig: verification_bounds}, we also report illustrative timing results. The average MILP solution time (including build time) for the 4-bus system was 21.01 ms, while a single Adam iteration (run on a single thread) required 53.12 $\mu$-seconds. Thus, if Adam terminates in $\sim$100 iterations (as it would in Panel 5), it would require 5.3 ms of solve time.}

\section{Conclusion}\label{sec: Conclusion}
Neural network (NN) verification is a very hard problem; but also crucial for building trust in machine learning methods for safety-critical systems. In this paper, we utilize and extend a GPU-accelerated tool ($\alpha,\beta$-CROWN) developed by the ML research community to solve verification problems which are pertinent to power systems. In contrast to NN verification problems in other fields, verifiers for power systems need to (i) be able to simultaneously verify for a large number of safety metrics (e.g. checking for violations of thousands of individual line flows at the same time), and (ii) be able to integrate physical constraints, such as the power flow equations. This paper introduces, for the first time to our knowledge, an exact transformation that converts the simultaneous verification problems to ReLU-based NN layers, which allows to be directly interpreted by verification solvers in a single shot. Our approach achieves $>$100x speedup compared to the Gurobi 10.0 solver. Second, this paper also introduces a mathematical framework for including equality and inequality constraints within the verification problem itself, which allow to verify that the NN output satisfies the power flow equations. Future work will seek to embed this developed framework within the actual source code of $\alpha,\beta$-CROWN.

\section*{Appendices}
\subsection*{A1. Integer Reformulation of min-min($\cdot$)}
Minimizing the smallest (or maximizing the largest) element of a set is a non-convex problem. The mixed integer reformulation is given via
\begin{subequations}\label{eq: non-convex_reformulation}
\begin{align}
\min_{x,y\in{\mathcal C}}\;\min(x,y)=\min_{x,y\in{\mathcal C}} & \quad t\\
{\rm s.t.} & \quad x\le t+b_{1}M\\
 & \quad y\le t+b_{2}M\\
 & \quad b_1+b_2=1\\
  & \quad b_1,b_2\in \{0,1\}.
\end{align}
\end{subequations}
where $b_1=0$ if $x$ is smallest, $b_2=0$ if $y$ is smallest, and $M$ is a sufficiently large constant.

\subsection*{A2. Dual Norm}
Given a norm $\left\Vert \cdot\right\Vert$, the \textbf{dual norm} of some vector $y$, denoted by $\left\Vert y\right\Vert _{*}$, is defined by
\begin{align}
\left\Vert y\right\Vert _{*}=\max_{\left\Vert x\right\Vert \le1}x^{T}y.
\end{align}
The $\ell_p$ and $\ell_q$ norms are each other's duals when $p$ and $q$ are Holder conjugates (i.e., $1/p+ 1/q=1$ and $p,q \in [1, \infty]$). Thus, quite fantastically, we have
\begin{align}\label{eq: pq_conj}
\left\Vert y\right\Vert _{q}=\max_{\left\Vert x\right\Vert _{p}\le1}x^{T}y,\forall p,q\in[1,\infty],\;\tfrac{1}{p}+\tfrac{1}{q}=1.
\end{align}
This property is exploited for solving the inner minimizations in \eqref{eq: ld_solve_soln} and \eqref{eq: ld_solve_soln_update}, where the $\min$ is swapped for $\max$ by simply negating the objective.

\subsection*{A3. Normalizing Input Bounds}
Consider the set of element-wise input bounds $\underline{x}\le x\le\overline{x}$. Defining the mean and half the range as
\begin{align}
x_{\mu} & =\tfrac{1}{2}(\underline{x}+\overline{x})\\
x_{\sigma} & =\tfrac{1}{2}(\overline{x}-\underline{x}),
\end{align}
we may normalize $x$ via
\begin{align}
\hat{x}=\tfrac{1}{x_{\sigma}}(x-x_{\mu}).
\end{align}
All elements of transformed variable $\hat{x}$ are now bounded by $\pm 1$, such that $-1\le\hat{x}\le1$. A unity infinity norm on $\hat{x}$ plus a linear inverse transformation (which can be encoded as a simple NN layer) now directly imply the original bounds on $x$:
\begin{align}
\left.\begin{array}{c}
\left\Vert \hat{x}\right\Vert _{p=\infty}\le1\\
\hat{x}\cdot x_{\sigma}+x_{\mu}=x
\end{array}\right\} \quad\Leftrightarrow\quad\underline{x}\le x\le\overline{x}.
\end{align}

\subsection*{A4. Collecting N-1 DC-OPF training data}
In order to train a NN to predict N-1 DC-OPF solutions, we collected a series of solutions to \eqref{eq: dcopfNm1}. For a given load vector $p_d$, \eqref{eq: dcopfNm1} minimizes generation costs subject to generation limits, power balance, and line flow limits; the line flow limits are additionally enforced for the loss of any single line. Using Gurobi 10.0, we randomly sample loads 1000 times and attempt to solve \eqref{eq: dcopfNm1} for each sample; infeasible load samples (i.e., samples for which no N-1 solution exists) are rejected. The resulting solution pairs $p^\star_d$, $p^\star_g$ are then used as NN training data points.
\begin{subequations}\label{eq: dcopfNm1}
\begin{align}
\min_{p_{g}}\quad & c_{g}^{T}p_{g}\\
{\rm s.t.}\quad & -\!M(1\!-\!b^{{\rm on}})\le p_{f}-Y_{b}E\theta\nonumber\\
 &\quad \quad \quad \quad \quad \quad \quad \quad\le M(1\!-\!b^{{\rm on}})\\
 & -\!Mb_{i}^{{\rm on}}\le p_{f,i}\le Mb_{i}^{{\rm on}}\\
 & p_{g}-p_{d}=E^{T}p_{f}\\
 & \underline{p}_{f}\le p_{f}\le\overline{p}_{f}\\
 & \underline{p}_{g}\le p_{g}\le\overline{p}_{g}\\
 & \sum b_{i}^{{\rm on}}\ge m-1\\
 & b_{i}^{{\rm on}}\in\{0,1\}.
\end{align}
\end{subequations}



\bibliographystyle{ieeetr}
\bibliography{references}


\end{document}